\documentclass[12pt]{article}
\usepackage{graphicx}        
\usepackage{amssymb}
\usepackage[numbers]{natbib}
\usepackage{bm}        
\usepackage{tikz}
\usepackage{smartdiagram}
\usepackage{algorithm2e}

\def\ra{\rightarrow}

\def\be{\begin{equation}}
\def\ee{\end{equation}}

\begin{document}

\title{Review of Parameter Tuning Methods for Nature-Inspired Algorithms}
\author{Geethu Joy$^{1,2}$, Christian Huyck$^{1}$, Xin-She Yang$^{1}$ \\[10pt]
1. School of Science and Technology, \\ Middlesex University London, \\ The Burroughs, London NW4 4BT, United Kingdom. \\[10pt]
2. Computer Engineering and Informatics, \\ Middlesex University Dubai, \\ Dubai Knowledge Park, P.O. Box 500697, \\ Dubai, United Arab Emirates. }

\date{}

\maketitle

\abstract{Almost all optimization algorithms have algorithm-dependent parameters, and the setting of such parameter values can largely influence the behaviour of the algorithm under consideration. Thus, proper parameter tuning should be carried out to ensure the algorithm used for optimization may perform well and can be sufficiently robust for solving different types of optimization problems. This chapter reviews some of the main methods for parameter tuning and then  highlights the important issues concerning the latest development in parameter tuning. A few open problems are also discussed with some recommendations for future research. }

\bigskip 


\bigskip
\noindent {\bf Citation Details:}
Geethu Joy, Christian Huyck, Xin-She Yang  in: {\it Benchmarks and Hybrid Algorithms in Optimization and Applications} (Edited by Xin-She Yang), Springer Tracts in Nature-Inspired Computing, pp. 33 -- 47 (2023). 
https://doi.org/10.1007/978-981-99-3970-1\_3
\\[15pt]

\section{Introduction}

Algorithms usually have algorithm-dependent parameters and the performance of an algorithm may largely depend on the setting of its parameter values. Therefore, parameter tuning becomes an important part of algorithm implementations and applications in practice~\citep{EibenSmit2011,YangHe2019,Yang2020nature,Talbi2009,Hussain2019Rev}.

Many design problems in engineering and industrial applications are optimization problems. To find the optimal solutions, or even sub-optimal solutions, to such optimization problems requires optimization techniques and algorithms. Nature-inspired algorithms are a class of recent optimization algorithms that are becoming popular and widely used in solving optimization problems. Like many other algorithms, each nature-inspired optimization algorithm usually has a few parameters that need to be properly tuned. Ideally, we should have a good tool to tune the parameters for a given algorithm so that it can maximize the performance of the algorithm under consideration. However, such tools do not exist. Therefore, tuning parameters can be a challenging problem, especially for tuning optimization algorithms. In essence, parameter tuning of an optimization algorithm is a hyper-optimization problem because it is the optimization of an optimization algorithm~\citep{Yang2013STA,YangBook2020,JoshiBans2020,Lacerda2021,RathoreFA2018,Lacerda2023}.

Even if a good tuning tool may exist so that we can tune an algorithm such that it performs well for a given problem, this tuned algorithm may not perform well for other problems, especially new problems or a different type of problems with unknown optimality. The reason is that the parameter-tuning process tends to use a given optimization problem (or a small set of optimization problems) to tune an algorithm, and consequently the tuned parameters may be problem-specific. If this is the case, algorithms have to be tuned again for every new problem, at least for every new type of problem. Therefore, parameter tuning can be a very time-consuming task.

Since parameter tuning can be computationally extensive, one naive way for solving optimization problems is probably to choose algorithms without any parameters (i.e., the so-called parameter-free or parameterless algorithms). However, such algorithms are rare in the context of optimization. In fact, many so-called parameter-free optimization algorithms in the literature are not entirely parameter free, and they still have some parameters, such as the population size of a swarm intelligence based algorithm. In many cases, researchers tend to mean that the number of key parameters are significantly reduced so that the algorithms require little or almost no tuning, even though users have to set the values of some hyper-parameters, such as the population size. But there is a serious issue with parameter-free algorithms because they are less flexible and less efficient.

For algorithms to be efficient for solving a class of optimization problems, the parameters in an algorithm should be fine-tuned so that the algorithm becomes effective for solving the specific class of problems under consideration. For an algorithm that is completely free from any parameters, if it does not work well for a given problem, there is no way that we can make it more efficient. In the end, we have to choose other algorithms to deal with the unsatisfactory solutions and try to find better solutions.

In addition to the above issues concerning parameter tuning, there is another problem, called parameter control. For parameter tuning, once an algorithm is properly tuned, the parameter values are then fixed when running the algorithm for solving problems. However, some studies suggested that it could be advantageous to vary the parameter values
during the execution or iterations of an algorithm. Rather than fixing the values of algorithm-dependent parameters, their values can be further tuned by varying or controlling how it may vary during the iteration. This type of problem is called parameter control.
At the present, it is not clear how to tune parameters effectively and how to control parameters properly for any given algorithm and a given set of problems. In fact, both are still open problems in this area.

Therefore, the purpose of this chapter is to provide  a timely summary of problems related to parameter tuning and parameter control. Some of the most recent studies concerning parameter tuning will be reviewed and discussed.

\section{Parameter Tuning}

To help our discussions about parameter tuning, let us use $A$ to represent an algorithm under consideration. The parameters in $A$ is a vector $p=(p_1, p_2, ..., p_m)$ of $m$ parameters. For a given problem $Q$, its optimal solution is denoted by $x_*$, which can be obtained by algorithm $A$ starting with a random initial solution (or any educated guess) $x_0$. Any solution at iteration $k$ is represented by a vector
$x^k=(x_1, x_2, ..., x_d)$ in the $d$-dimensional space.

The iterative solution process of obtaining the optimal solution to problem $Q$ can be represented by
\be x^{k+1}=A(x^k, p, Q). \ee
In the limit of $k \ra \infty$, we have $x^{k+1} \ra x_*$ as the optimal solution or the best solution that the algorithm can find.

\subsection{Schematic Representation of Parameter Tuning}

Since the parameter vector $p$ can take different values. Initially, we may have $p=p_0$, but $p_0$ can be obtained by random initialization or an educated guess.

Suppose we have a performance measure for algorithm $A$, which can be denoted by $\mu$. For example, $\mu$ can be considered as the success rate or $1/T$ where $T$ is the time needed to find the optimal solution to $Q$.

With such notations, parameter tuning can be represented as
an optimization problem to find $p$ so that
\be \max \;\; \mu=F(A, p, Q), \ee
where $F()$ is a known function to be defined by the user.
However, this requires an iterative process
\be p_*=\lim_{j \ra \infty} \arg_p F(A, p^j, Q). \ee
In practice, we cannot afford a large number of iterations for parameter tuning. Thus, we typically set a computational budget $B$ for parameter tuning, and $B$ can be considered as the maximum number of tuning iterations. Here, $p_*$ is the vector of tuned parameter values.

\subsection{Different Types of Optimality}

Using the above notations, for a given problem $Q$ and a selected algorithm $A$, there are two types of optimality: the optimal solution $x_*$ and the optimal setting $p_*$.

Therefore, parameter tuning involves two optimization problems at the same time. One optimality does not lead to the other optimality.

Traditionally, most researchers focus on the first optimality of $x_*$ so that they can obtain the optimal solution, or sub-optimal solutions, to their optimization problems. This is fine for small-scale problems or the optimization process is not too time-consuming. For complex, large-scale problems, tuning of parameters becomes essential to save computational costs. A well-tuned algorithm may reduce the computational time significantly, though the tuning of algorithm-dependent parameters is an art at the moment, rather than a systematic methodology.

\subsection{Approaches to Parameter Tuning}

\begin{figure}
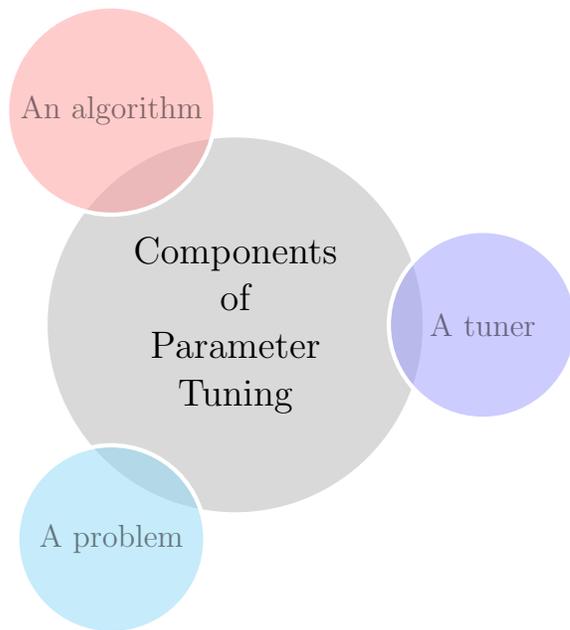

\begin{center}
\smartdiagramset{planet text width=2.5cm}
\smartdiagram[bubble diagram]{Components \\ of \\ Parameter \\ Tuning, An algorithm, A problem, A tuner}
\caption{Three components of parameter tuning. \label{Fig-Para-100}}
\end{center}
\end{figure}

Based on the two different types of optimality, there are different ways for approaching parameter tuning~\citep{YangBook2020}.
For parameter tuning and control in general, there are always three components: an algorithm, a tuning tool, and a problem (see Fig.~\ref{Fig-Para-100}).

Obviously, the first key component is the algorithm to be tuned. To tune the algorithm, we have to use a tuning tool (i.e., a tuner), and apply the algorithm to a problem or a problem instance so as to evaluate the performance of the algorithm being tuned for solving the problem. Therefore, it can be expected that the tuned results in terms of parameter settings can be both algorithm-specific and problem-specific.

In addition to parameter tuning, the variations of parameters during iterations can be also advantageous, and such variations are often called parameter control~\cite{Eiben1999,YangHe2019}. Though our emphasis here is mainly parameter tuning, we will also consider parameter control when appropriate.

\subsubsection{Hyper-Optimization}

Since parameter tuning is the optimization of an optimization algorithm, we can refer it to as hyper-optimization~\cite{Yang2013STA}. Some researchers
also called it meta-optimization in a slightly different context~\cite{Skakov2018}. In terms of achieving optimal tuning, we can have two different structures: sequential and parallel.

\begin{figure}[h]
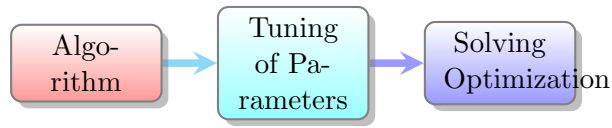
 \begin{center}
\smartdiagramset{back arrow disabled=true}
\smartdiagram[flow diagram:horizontal]{Algorithm, Tuning of Parameters, Solving \\ Optimization}
\caption{Sequential structure of parameter tuning. This is also called offline tuning. \label{Fig-100}}
\end{center} \end{figure}

    The sequential structure means that we tune the algorithm $A$ first for the given problem $Q$ so as to find the optimal setting of parameters $p_*$. Whether it is achievable or not
    is a separate issue. For the moment, suppose that it is possible to find $p_*$ in this way. Then, we apply the optimal $p_*$ for
    algorithm $A$ to solve $Q$, and then we obtain $x_*$ as the optimal solution. Again, whether this is achievable or not is another issue. This structure can be schematically represented in Fig.~\ref{Fig-100}.

\begin{figure}[h]
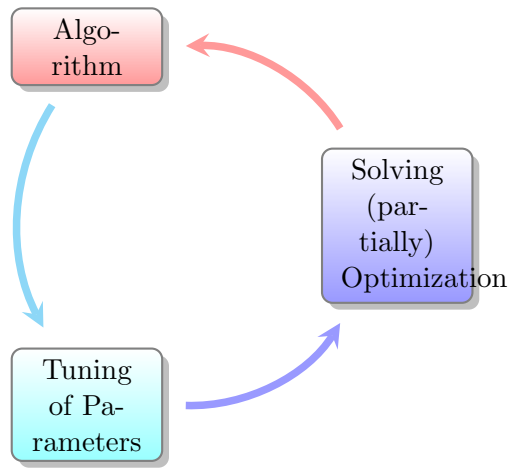
 \begin{center}
\smartdiagram[circular diagram:horizontal]{Algorithm, Tuning of Parameters, Solving (partially) \\ Optimization}
\caption{Loop structure of parameter tuning. This is sometimes called online tuning. \label{Fig-200}}
\end{center} \end{figure}

Another way for parameter tuning is to use a parallel structure or a loop structure to carry out tuning and problem-solving iteratively. This can be represented as steps in Fig.~\ref{Fig-200}. If the parameter values can be changed again inside the step of solving the optimization problem, it becomes a problem of parameter control.

\subsubsection{Multi-objective Approach}

It is difficult to justify which structure is better. In fact, existing studies do not provide any evidence whether these structures can work effectively.

Another way of looking at this parameter-tuning problem is to carry out both optimization processes simultaneously, which can essentially lead to a bi-objective optimization problem
\be \min \;\; Q=f(x) \quad \textrm{ and } \quad  \max \quad \mu. \ee
From this multi-objective perspective, we know from the theory of multi-objective optimization that there is no simple single optimality.
Instead, we should have multiple, equally good solutions, which will form the Pareto front. In this sense, we can only find Pareto-optimal settings. This means that there is no unique parameter setting in general. If the problem $Q$ changes to a different problem, the tuned parameters should also change so as to maximize both objectives.

\subsubsection{Self-Tuning Approach}
Another possibility is to combine the tuning process with the optimization problem to be solved. This leads to the self-tuning framework, developed  by Yang et al.~\citep{Yang2013STA}.

The main idea is to extend the solution vector $x=(x_1, x_2, ..., x_d)$ to include $p=(p_1, p_2, ..., p_m)$ so as to form a vector of $d+m$ dimensions
\be u=(x, p)=(x_1, x_2, ..., x_d, p_1, p_2, ..., p_m), \ee
we then use $u$ to solve both problems to find the optimal $x_*$ and $p_*$. In this case, we need to combine the bi-objective functions to form a composite objective function
\be \min \;\; g=\alpha f(x)- \beta \mu,  \ee
to be minimized. Here, $\alpha>0$ and $\beta>0$ are two weights.

There are other possibilities or approaches for parameter tuning. For example, we can consider parameter tuning as a self-organized system~\cite{Keller2009}.
So, in the rest of chapter, we will review some of the recent studies.

\section{Review of Parameter Tuning Methods}

\subsection{Generic Methods for Parameter Tuning}

The classifications of parameter tuning methods can be difficult, depending on the perspective of the classification~\cite{YangBook2020,Phan2020,Huang2020}. For example, Huang et al.~\cite{Huang2020} used the action of generation of parameters to divide into: simple, iterative or high level generation-evaluation schemes. However, this categorization does not provide enough details about the mechanisms of parameter tuning. Here, we will loosely divide parameter tuning methods into eight different categories:

\begin{enumerate}
\item \emph{Manual or brute force method}. A naive way to tune parameters of a given algorithm is to try every possible combination. In general, this is not possible and the most time-consuming method. However, if there is only one parameter, we can subdivide the parameter into smaller discrete intervals, then we can potentially try every possible combination of intervals. This allows us to find the best parameter intervals and enable further fine-tuning if needed. The advantage of this method is that it will ensure to explore every possible interval. However, the major disadvantage is the high computational cost. Thus, this method is only suitable for tuning a single parameter in a discrete range.

\item \emph{Tuning by systematic scanning}. A slightly improved version of brute force tuning is the systematical scanning. Suppose a parameter $p$ varies in the range of $[a, b]$, we can split this interval into
    multiple (not necessarily equal) subintervals, and carry out the scanning systematically. For example, if $p$ varies in [0,10], we can try $p=0.0, 0.1, 0.2$, ..., $9.8, 9.9$ and $1.0$. Suppose that we find that the best possible range or values is (for example) 2.5, then we can fine-tune it from 2.4 to 2.6 by using $p=2.40, 2.41, 2.42$, ..., $2.59$ and $2.60$. This process continues until a predefined stop criterion is met.

\item \emph{Empirical tuning as parametric studies}. In practice, most researchers would use some sort of empirical tuning. Empirical tuning of parameters starts with either an educated guess or a known value from an existing study, then perturbs the parameter values around a larger range so as to fine tune the parameter or test the robustness as well as the sensitivity of the parameter under consideration. For multiple parameters, we can apply the same procedure to do such parametric studies. The advantage of this method is that it is usually quick to obtain useful parameter values, but it is difficult to provide any useful insights.

\item \emph{Monte Carlo based method}. The Monte Carlo based method is a statistical sampling method with relatively solid theoretical basis and the error analysis to ensure that such parameter tuning can work well in practice. However, due to the slow convergence of Monte Carlo based methods, a large number of samples (typically thousands) or parameter settings need to be evaluated, which is still time consuming. But compared with manual or brute force tuning, this is a major improvement and can work well in practice. A possible further improvement is to use the so-called quasi-Monte Carlo method with low-discrepancy sequences, which may reduce some of the computational efforts, though we have not seen any significant work in this area yet.

\item \emph{Tuning by design of experiments}. As parameter tuning tends to be time-consuming, design of experiments will be useful for tuning parameters with a fixed computational budget. For example, for multiple parameter tuning with a fixed number of possible tuning experiments, methods via design of experiments  such as the Latin hypercube and factorial techniques can provide a good approach, though there is no guarantee that a good set of parameter settings can be achieved. Obviously, if the tuning budget increases, better parameter settings are more likely to be achievable.

\item \emph{Machine learning based methods}. Strictly speaking, machine learning based algorithms or methods are usually more sophisticated than metaheuristic algorithms. To use machine learning based methods seem to be a bit over the top and make things more difficult to understand. However, some studies have shown that they can work well for tuning parameters. But this requires good data and more computational efforts. Therefore, this should not be a first choice as a method for tuning algorithm-dependent parameters.

\item \emph{Adaptive and automation methods}. In recent years, adaptive parameter tuning becomes more popular due to its ease of implementation. Self-adaptive methods seem to be more effective in tuning parameters online. However, the design of adaptive rules can be another unresolved challenging problem. In addition, automation methods tend to make adaptive tuning more effective and easier to implement. Again, the design of such automation itself is a difficult task.  As an example, the self-tuning framework can be considered as a kind of automation method~\cite{Yang2013STA}.

\item \emph{Other methods}. There are other methods that we may not be able to put into the above categories. For example, parameter tuning by sequential optimization is an interesting method~\cite{Trindade2019}, which may not fit into any of the above categories. In addition, hybrid approaches to parameter tuning~\citep{Shadkam2021}, and multi-objective approach to parameter tuning can be considered as alternative methods, which can be equally effective~\cite{Talbi2013}. However, their computational efforts may vary significantly. Also, there is an emerging literature about parameter tuning, which requires a more systematical review in the near future.

\end{enumerate}

Though the above categorization include almost all known parameter tuning methods for offline tuning,   online tuning may be relevant to some of these categories. More specially, adaptive and self-adaptive tuning, Monte Carlo based tuning, machine learning based tuning, and other methods are relevant to online tuning.

\subsection{Online and Offline Tuning}

Though the theoretical understanding of parameter tuning and parameter control may be quite limited, the numerical experiments on parameter settings are relatively extensive. Here, in the rest of this chapter, we will focus on the review of some of these studies.

Loosely speaking, the fine-tuning of parameters can sometimes increase the efficiency of algorithms if the tuning can help to reach a better balance of exploration and exploitation, thus providing a better trade-off and a potentially higher rate of convergence.  In general, parameter settings can influence both the quality of the solutions and the balance between local and global exploration. For example, it has been demonstrated using different variants of PSO that the best parameter values or algorithm configurations can be time dependent~\cite{Harrison2019}. In addition, performance can be enhanced if exploitation and exploration can be more balanced~\citep{Sababha2018}.

According to~\cite{Hutter2007AutomaticAC}, parametrization is described as  finding the best settings of parameters for a given optimization technique by solving a problem instance so as to improve the known performance index. Thus, the process of tuning or setting parameters of a metaheuristic algorithm is also referred to as algorithm configuration in the literature~\cite{Eryoldas2022}.

The term offline parameter tuning or parametrization refers to the setting of parameters prior to the execution of the metaheuristic algorithm used for solving a set of problems. In contrast, online parameter tuning or parametrization means that the parameters can be updated while the metaheuristic algorithm is being executed~\cite{Biratt2009}.

Typically, offline tuning methods are carried out in the early stage, usually as part of the pre-processing phase. The values of the parameters are set by observing the parameter values used in similar experiments or setup. Sometimes, additional insights from statistical analysis of the algorithm's performance on similar problems, while deciding the parameter values during offline tuning, can ensure a decent performance of the algorithm~\cite{inproceedings3}. In this sense, offline parametrization can in general be divided into four categories: manual parameterization, parameterization by analogy, parameterization by design of experiments (DOE), and search based parameter tuning. Offline parameter tuning methods can usually require significant computational efforts to get some sensible statistical measures, and such methods include the sequential model-based optimization~\cite{HutterH2011}, ParamILS~\cite{Hoos2012}, F-Race~\cite{Birattari2010} and DOE~\cite{bartz2007experimental}.

On the other hand, online parameter control methods usually require the minimal user intervention and absence of preliminary experimentation. However, there is no guarantee that computational efforts may be less.
Typically, parameter control (online tuning) is not part of pre-processing, and in essence is an {\it ad hoc} process. The values of  parameters may be adjusted in each iteration or after a fixed number of iterations, according to the feedback or outputs obtained from the previous runs. In this sense, online parameter setting methods are more specific and may potentially provide a better performance for the algorithm in terms of its efficiency and effectiveness. Since such tuning is problem-specific, the method in this case can be implemented as part of the solver~\cite{inbook2}. However, the disadvantages are over-specialization of this kind of parametrization method to the given algorithm and a given type of problems. In addition, the introduction of new parameters as part of the tuning process itself may expand the parameter domain significantly. In general, the performance of a parameter setting method is evaluated based on the speed of the parameterized algorithm in converging to the solution for a problem set or instances and the quality of the solution or a combination of both~\cite{Yasemin2022}.

Some authors suggested that online parametrization method could use an approximate gradient search with line search to search for parameter values in the parameter domain~\cite{article4}. The efficiency of this method was demonstrated using differential evolution with two test suites. This approach was further enhanced by a grid based search~\cite{article5}.
The search starts with some arbitrary values for the parameters. The neighboring values of the initial arbitrary values are then selected using grid search and the algorithm's performance for the newly selected parameter values is estimated using short runs. The current set of parameter values are then used for several iterations before a new set of values are selected using a grid search method. Then, their influence on the algorithm's performance is evaluated using more runs.

For online tuning methods, an eTuner was developed by~\cite{inproceedings1}. They claimed that their eTuner method can overcome the limitations of some naive brute force strategy.
		
Though it is no surprising that  evolutionary algorithms are used to tune the parameters of a metaheuristic algorithm~\cite{Skakov2018}, though the evolutionary algorithms themselves are also needed to be tuned. Imagine a case that the evolutionary algorithms are well tuned, then these tuned algorithms can be used to tune metaheuristic algorithms. Although it is still  a very computationally intensive process, certain automation reduces the workload involved and the efficiency of this method is demonstrated using facility location problem~\cite{Skakov2018}.

Some researchers referred to the process of using a metaheuristic algorithm to tune the parameters of a metaheuristic algorithm as meta-optimization, whereas other researchers called it hyper-optimization. It is worth pointing out that terminologies in this area are not consistent yet and there are no agreed sets of common terminologies. However, we are trying to use a consistent set of terminologies in this chapter.

\subsection{Self-Parametrization and Fuzzy Methods}

A self-parametrization method was developed by the authors of \cite{Santos2022}, which used a metaheuristic with the parameter space and the solution space. However, it is worth pointing out that  this self-parametrization method is not a self-tuning method.
A prototype of self-parametrization was then used to analyze the performance of the proposed parametrization framework. For the tested scheduling problem and the traveling salesman problem, they showed that their method was statistically significant better than some manual-parameter-tuning methods.

A method using fuzzy inference was suggested in \cite{article10} to tune the grey wolf optimizer. A tuning method that combines both the fuzzy C-means clustering~\cite{article23} and Latin Hypercube sampling~\cite{article24} was proposed in \cite{Yasemin2022}.
The authors suggested early elimination of parameter configurations to reduce computation cost and tuning time. Each tuning algorithm has its own advantages and disadvantages.

An evaluation method based on Monte-Carlo simulation was proposed in~\cite{article37} to explore the influence of parameters on the performance of metaheuristic algorithms. In addition, some of the major pitfalls in algorithm configuration and the best practices to avoid or overcome these pitfalls have been discussed in~\cite{article33}.

\subsection{Machine Learning Based Methods}

Some researchers used reinforcement learning for tuning parameters. Though a different approach and from a different perspective, it can involve more computational efforts. In general, parameter tuning methods involving the use of reinforcement learning has the following drawbacks~\cite{Lacerda2023}:
computationally intensive, final results can be highly dependent on the hyper-parameters in learning, and very limited number of benchmark functions.

In the current literature, other machine learning algorithms were also used to tune the parameters of metaheuristic algorithms~\cite{Tan2021,article43}. In \cite{Tan2021}, a random forest was used to tune the parameters of bee colony optimization. The improvement in efficiency of the tuned version was demonstrated on some instances of the Traveling Salesman Problem. In \cite{article43}, the authors used a set of machine learning techniques to tune the parameters of PSO. In addition, a meta-learning approach was developed for tuning parameters~\cite{Hekmat2019}, which seemed to be effective. Randomization techniques can also be very effective for hyper-parameter optimization~\cite{Bergstra2012}, if hyper-parameter optimization can be considered as a kind of parameter tuning.
Similarly, hyper-parameters can be tuned using deep neural networks~\cite{Yoo2019}, whereas a statistical learning based approach was developed for parameter fine-tuning of metaheuristics~\cite{Calvet2016}.

There may be other tuning methods that we have not covered in this chapter, as the literature is evolving in this area. In the rest of this chapter, we will discuss some of the key issues in parameter tuning and will make some recommendations for future research.

\section{Discussions and Recommendations}

Based on the brief review in this chapter, it seems that there have been a lot of efforts in investigating parameter tuning methods, either offline or online. However, it still lacks some true insights about how these parameter tuning methods may work. Equally, in most cases, these methods may not work well as expected, there are no insights or indicators to explain the possible reasons. Most important, there are some serious questions and open problems related to parameter tuning and parameter control.

Based on our observations and analysis, there are three main issues concerning parameter tuning: non-universality, high computational efforts and lack of insights into how the algorithms may behave in terms of  convergence.

\begin{itemize}
\item Non-universality: Almost all the parameter tuning methods require three components: an algorithm, a problem or problem instance, and a tuning tool/rule. Therefore, the settings of parameters that have been properly tuned (if they are achievable) are both problem-specific and algorithm-specific. Thus, we cannot expect the tuned parameter settings that work for an algorithm for a set of problems can be equally effective  for a different set of problems. In fact, there is no universality in parameter tuning, and further tuning seems to be needed for new problems.

\item High computational efforts: The main barrier for parameter tuning and parameter control is the high computational cost needed for tuning parameters properly. Multiple runs of the algorithm to be tuned with multiple parameter combinations are needed. Thus, it is highly needed to develop new tuning methods so as to minimize or at least reduce the overall computational efforts significantly.

\item Lack of insights: Though a good spectrum of diverse tuning methods exist in the current literature, almost all these methods are heuristic methods to  a certain degree. Even the tuning methods may work quite well, it lacks insights how they work and under what conditions. An even more important problem is that if the parameters are best tuned, what do they imply in practice? Alternatively, how good parameter settings may influence the convergence of the algorithm and consequently its convergence is still an open problem.

\end{itemize}

With the above issues and open problems, we would like to make some recommendations for future research concerning parameter tuning in the following three areas: theory, link to convergence analysis,
and universal tuning.

\begin{itemize}
\item \emph{Theoretical analysis}. It is highly needed to carry out some mathematical or theoretical analysis of how parameter tuning works. Ideally, the possible ranges of parameters can be derived or estimated from analyzing the algorithm itself.

\item \emph{Relationship to algorithm convergence}. It can be expected that properly tuned algorithms should have better convergence. However, it is not clear how the parameter settings can directly affect the convergence rate of the algorithm under consideration.
    Therefore, it is desirable to explore the possibility of establishing some theoretical link between parameter settings and convergence rates. This is still an open problem and more research is highly needed.

\item \emph{Universal tuning}. As we have seen earlier, parameter tuning methods seem to produce results that are both algorithm-specific and problem-specific, which limits the usefulness of the tuned algorithms. Ideally, the tuning should be a black-box type without the knowledge of the algorithm and problems, and the tuning can be more general and the results can be potentially universal. However, it is not clear if such methods are possible and how to design such methods. Future research should explore more in this area so as to develop more generic tuning tools.

\end{itemize}

In addition to the two types of optimality we briefly explained earlier, there is a third optimality that concerns the optimal rate of convergence. When we say the parameters are best tuned, it all depends on the performance metric used. If the metric is to minimize the number of iterations or increase of the accuracy of the algorithm for solving a given set of problems, this is not exactly the same as the best rate of convergence. Thus, there are some subtle differences in what we mean `best settings'. In the current literature, different researchers seem to use the best settings to mean different things, according to different rules/metrics. This issue needs to be resolved and a set of useful rules should be agreed upon before we can truly make insightful understanding in parameter tuning and control.

As we have seen in this chapter, there are many issues and open problems concerning parameter tuning and parameter control. We sincerely hope that this brief review can inspire more research in this area in the near future.

\end{document}